\documentclass[journal]{IEEEtran}

\usepackage{cite}

\ifCLASSINFOpdf
   \usepackage[pdftex]{graphicx}
   \DeclareGraphicsExtensions{.pdf,.jpeg,.png}
\else
   \usepackage[dvips]{graphicx}
   \DeclareGraphicsExtensions{.eps}
\fi

\usepackage{amsmath,amssymb}
\usepackage[linesnumbered,ruled]{algorithm2e} 
\usepackage{algorithmic}
\usepackage{array}

\ifCLASSOPTIONcompsoc
  \usepackage[caption=false,font=normalsize,labelfont=sf,textfont=sf]{subfig}
\else
  \usepackage[caption=false,font=footnotesize]{subfig}
\fi

\usepackage{url}
\usepackage[usenames]{color}
\usepackage[table]{xcolor}
\usepackage{multirow}
\usepackage{diagbox}
\usepackage{threeparttable}
\newcolumntype{M}[1]{m{#1}<{\centering}}
\usepackage{setspace}
\usepackage{arydshln}

\newcommand{\ie}{\textit{i}.\textit{e}.}
\newcommand{\eg}{\textit{e}.\textit{g}.}
\usepackage[pagebackref=true,breaklinks=true,linkcolor=red,anchorcolor=blue, citecolor=green,colorlinks,bookmarks=false]{hyperref}

\makeatletter
\def\endthebibliography{%
  \def\@noitemerr{\@latex@warning{Empty `thebibliography' environment}}%
  \endlist
}
\makeatother

\begin{document}

\title{Temporal Self-Ensembling Teacher for Semi-Supervised Object Detection}

\author{Cong~Chen$^\dagger$,~
        	  Shouyang~Dong$^\dagger$,~
        	  Ye~Tian,~
        	  Kunlin~Cao,~ \\
        	  Li~Liu,~\IEEEmembership{Senior Member,~IEEE} and Yuanhao~Guo$^\star$,~\IEEEmembership{Member,~IEEE}
\thanks{C. Chen and K. Cao are with Keya Medical Technology, ShenZhen, 518116.} 
\thanks{S. Dong is with Software Department at Cambricon, Beijing, 100010.}
\thanks{Y. Tian is with Hippocrates Research Lab at Tencent, Shenzhen, 518052.}
\thanks{Li Liu is with the Colloge of System Engineering, National University of Defense Technology, China and is also with Center for Machine Vision and Signal analysis at the University of Oulu, Finland.}
\thanks{Email: li.liu@oulu.fi}
\thanks{Yuanhao Guo is with Institute of Automation, Chinese Academy of Sciences, Beijing, 100190.}
\thanks{Email:yuanhao.guo@ia.ac.cn}
\thanks{$^\dagger$ S. Dong and C. Chen are equal contributors.}
\thanks{$^\star$ Y. Guo is the correspondence author}}

\markboth{Preprint to journal}%
{Chen \MakeLowercase{\textit{et al.}}: Temporal Self-Ensemble Teacher for Semi-Supervised Object Detection}

\maketitle

\begin{abstract}
This paper focuses on the problem of Semi-Supervised Object Detection (SSOD). Recently, Knowledge Distillation (KD) has been widely used for semi-supervised image classification. However, an empirical adoption of these methods for SSOD has the following obstacles. (1) The teacher model serves a dual role as a teacher and a student, such that the teacher predictions on unlabeled images may be very close to those of student, which limits the upper-bound of the student. (2) The extreme foreground-background class imbalance issue existing during training of dense detectors hinders an efficient knowledge transfer from teacher to student. To address these problems, in this work, we propose a novel framework called Temporal Self-Ensembling Teacher (TSE-T) for SSOD. Differently from the conventional KD based methods which keep the teacher constant, we devise a temporally evolved teacher model. First, our teacher model ensembles its temporal predictions for unlabeled images under stochastic perturbations. Such data augmentation and temporal ensembling strategy increase data diversity, which thus improves prediction accuracy. Second, our teacher model ensembles its temporal model weights with the student model weights by an exponential moving average (EMA) which allows the teacher gradually learn from the student, yielding temporally diverse teacher model. These self-ensembling strategies collaboratively lead to better teacher predictions for unlabeled images. Finally, we use focal loss to formulate the consistency regularization term to handle the data imbalance problem in SSOD, which is a more efficient manner to utilize the useful information from unlabeled images than a simple hard-thresholding strategy which solely preserves confident predictions. Evaluated on the widely used VOC and COCO benchmarks, the mAP of our method has achieved $80.73\%$ and $40.52\%$ on the VOC2007 test set and the COCO2014 \textit{minval5k} set respectively, which outperforms a strong  fully-supervised detector by $2.37\%$ and $1.49\%$. Furthermore, our method sets the new state-of-the-art in SSOD on VOC2007 test set which outperforms the baseline SSOD method by $1.44\%$. The source code of this work is publicly available at \url{http://github.com/syangdong/tse-t}.
\end{abstract}


% \fi
\IEEEpeerreviewmaketitle

\begin{figure}[!t]
  \centering 
  \includegraphics[width=0.5\textwidth]{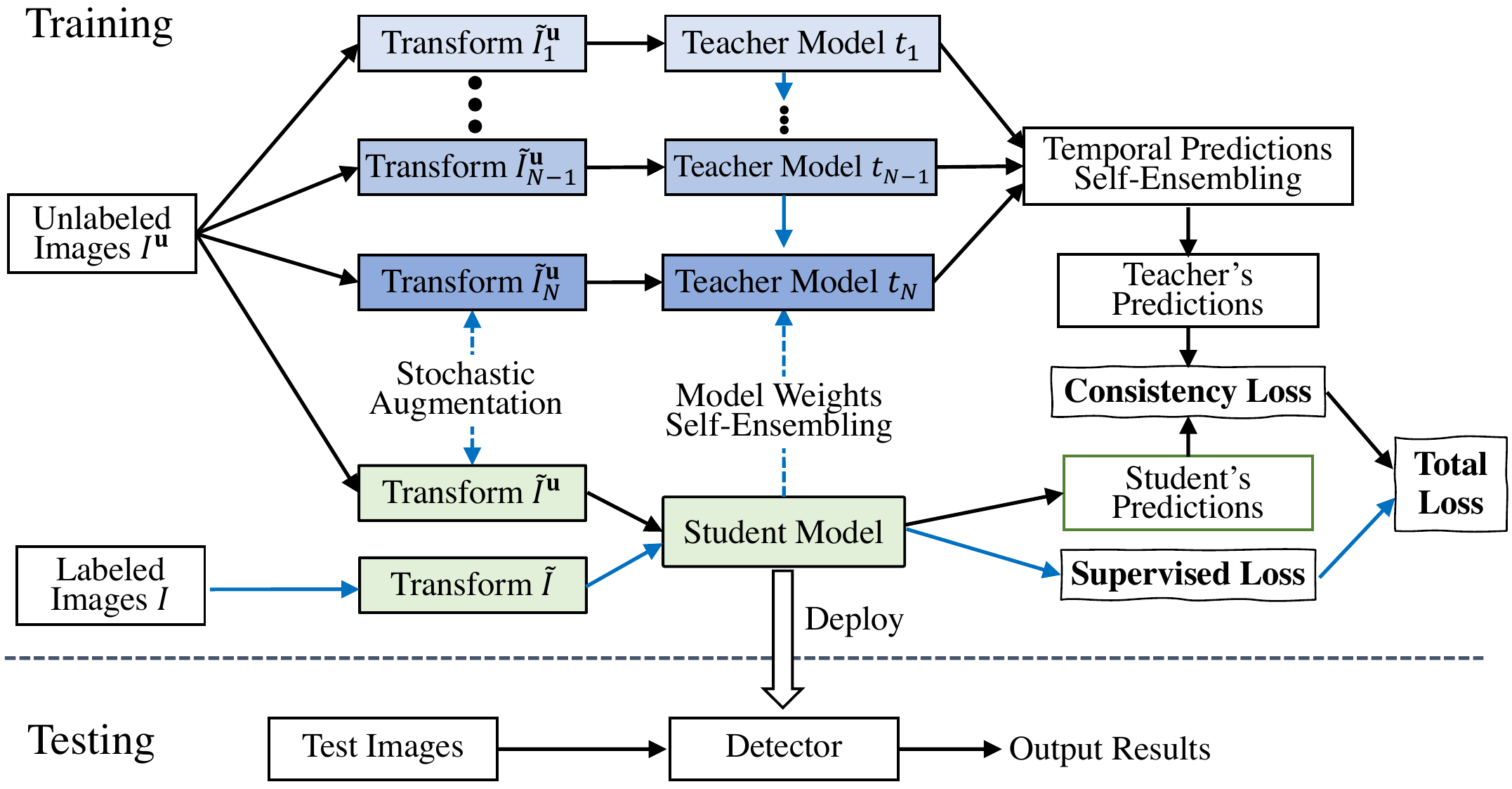}
  \caption{\textbf{The Framework of the Proposed TSE-T Model for SSOD.} At training time, unlabeled images under stochastic transformations like random horizontal flip are predicted by a series of consecutive teacher models. The results are then aligned and ensembled to obtain the teacher predictions which are subsequently used as targets to regularize the training of the student model. We keep the teacher model evolved using an EMA which results in temporally diverse teacher model. At testing time, the trained student model is deployed for object detection for unseen images.}
\label{fig::fig1}
\end{figure}

\section{Introduction}
\label{sec::intro}

Object detection is the cornerstone of computer vision, as many high level vision tasks fundamentally rely on the ability to recognize and localize visual objects. Object detection thus touches many areas of artificial intelligence and information retrieval, such as image search, data mining, question answering, autonomous driving, medical diagnosis, robotics and many others \cite{liu2020deep, masi2018deep, zheng2016person, chen2017multi, mckinney2020international}. The recent resurgence of interest in artificial neural networks, in particular deep learning, has tremendously advanced the field of generic object detection, and in the past few years a large number of detectors \cite{ren2015faster,lin2017focal,he2017mask,redmon2016you,liu2016ssd,law2018cornernet,cai2018cascade,liu2020deep} have sprung up to improve the detection performance from some aspects like accuracy, efficiency or robustness.

Current state-of-the-art detectors \cite{ren2015faster,lin2017focal,he2017mask,liu2016ssd,cai2018cascade} are learned in a fully supervised fashion, which requires large scale labeled data with many high quality object bounding box annotations or even segmentation masks. Gathering bounding box annotations or segment masks for every object instance is time consuming and expensive, especially when the training dataset contains a huge number of images or even videos, as it requires intensive efforts of experienced human annotators or experts (\eg, medical image annotation) \cite{everingham2010pascal,lin2014microsoft,ILSVRC15, DBLP:journals/corr/abs-1811-00982, liu2020deep}. Furthermore, manual bounding box/segmentation mask labeling may introduce a certain amount of subjective bias. In addition, the generalizability of fully supervised detectors is limited. By contrast, there are massive amounts of unlabeled images which are acknowledged valuable, and the key is how to make good use of them \cite{tarvainen2017mean,french2017self,jeong2019consistency,radosavovic2018data,he2019momentum,kolesnikov2019revisiting,kolesnikov2019revisiting,goyal2019scaling,doersch2017multi}.

The time consuming and expensive annotation process of accurate bounding boxes of object instances is sidestepped in Weakly Supervised Object Detection (WSOD), which only utilizes image level annotations that show the presence of instances of an object category \cite{oquab2015object,wan2018min,zhang2019leveraging}. WSOD methods may achieve a relatively good performance if provided with a large number of image level annotations, however the performance is hardly competitive to their fully supervised counterparts. Considering a generic situation in object detection, we have a limited number of labeled images \cite{everingham2010pascal,lin2014microsoft}, but a huge number of unlabeld images (\eg, the massive amounts of unlabeled image available from the Internet). Thus, Semi-Supervised Learning (SSL), which falls between supervised and unsupervised learning, has shown promising results to reduce the gap between. SSL has been extensively studied in image classification problem \cite{zhu2005semi,zhu2009introduction,chapelle2009semi,lee2013pseudo}, while it has received significantly less attention in object detection. In this work, our main focus is SSL for object detection.

Classical deep learning based SSL methods use the maximum predictions for unlabeled images as pseudo labels to improve the classification performance of the neural networks \cite{lee2013pseudo}. The recently developed Knowledge Distillation (KD) \cite{hinton2015distilling,phuong2019towards} aims at training a light weight student model regularized by a cumbersome teacher model, which was originally used for deep model compression but later widely used to solve SSL problems. Quite a few KD based SSL methods have been proposed \cite{rasmus2015semi,sajjadi2016regularization,laine2016temporal,tarvainen2017mean}, and the key to these methods is to construct a well-performed teacher to obtain stable and reliable predictions when giving unlabeled images during training. The teacher predictions for unlabeled images can be used as targets (well-posed logits or soft labels) to regularize the training of the student in order to obtain similar predictions on the unlabeled images, yielding a well-trained student to approach the performance of the teacher. This can be implemented by using the consistency regularization between the teacher and student predictions which routinely takes the form of Mean Squared Error (MSE) loss.

So far, however, only a limited number of works have applied similar ideas in a more challenging task, like SSOD \cite{jeong2019consistency, tang2020proposal}. The main challenges are as follows. (1) The teacher model in these KD based methods often serves a dual role as a teacher and a student. In image classification, it is sufficient to solely handle a unique prediction per image, but for the object detection which is a more complicated task to identify objects category and localize them simultaneously, such teacher model may produce very close predictions as the student. The risk behind is that the performance improvement of the student may be limited using unlabeled images. (2) The predictions in object detection are rather dense during training because an object is probable to present at every location in an image and an image may contain multiple objects. This issue is easy to handle in supervised object detection because a unique ground-truth is provided. However, this is difficult to tackle for SSOD because the teacher predictions acts the role to provide "annotations" for student model and these "annotations" may lead to severe data imbalance problem. Therefore, a direct adoption of the widely used consistency regularization term from SSL to SSOD is hampered. A recent method named Consistency based Semi-supervised Learning for Object Detection (CSD) \cite{jeong2019consistency} tackles this problem by simply thresholding out the low confident predictions. There are several limitations of this work. Given unlabeled images, (1) the teacher and student are identical which may result in similar predictions, and (2) the simple thresholding-out strategy may ignore useful information.

In this work, we aim at a simple but generic solution to alleviate the above issues, further improving the SSOD. To this end, we propose the \textbf{\textit{T}}emporal \textbf{\textit{S}}elf-\textbf{\textit{E}}nsembling \textbf{\textit{T}}eacher model, coined \textbf{\textit{TSE-T}}. We show the framework of our method in Fig.\ref{fig::fig1}. TSE-T model is devised on top of the KD framework which consists of a teacher and a student model. Both the teacher and student are initiated from a pre-trained detection network using fully-supervised manner. At semi-supervised training time, the teacher obtains the predictions for both category and localization of all possible objects presenting in the unlabeled images. The student also obtains its detections for these unlabeled images. The KD framework aims to minimize the dissimilarity between teacher and student predictions, which is implemented by using the consistency regularization between them. At testing time, the trained student model is deployed for the object detection in unseen images.

Based on the above base framework, our TSE-T model proposes the following novelties. 

(1) Our first goal is to enhance the performance of the teacher model on object detection in unlabeled images. To this end, we devise a  temporally updated teacher model which is asynchronous from the training of student.

Specifically, instead of using a constant teacher as proposed in the original KD based methods \cite{hinton2015distilling, radosavovic2018data}, our TSE-T model devises a teacher which ensembles its temporal predictions from consecutive training epochs for the unlabeled images under stochastic perturbations (random transformations like horizontal flip). This type of data augmentation and temporal predictions ensembling strategy has been widely used to effectively improve the prediction accuracy in SSL \cite{laine2016temporal}. Moreover, our teacher model ensembles its temporal model weights with the student model weights which allows the teacher to gradually learn from the student. The evolution of teacher model is decoupled from the training of student, which somehow prevents the teacher obtaining similar predictions from the student. 

These self-ensembling strategies together raise data and model diversity, thus yielding stable and reliable teacher predictions for unlabeled images which can be consequently used as better targets to train the student. The proposed TSE-T model substantially distils knowledge of multiple image geometric transformations from a well-trained teacher to the student. On the other words, the student is guided to imitate the behavior of teacher by its predictions on unlabeled images implemented in the form of a consistency constraint, thus leading the student's performance to approach the teacher. 

(2) Our second goal is to solve the data imbalance problem but avoid using hard-thresholding. A hard-thresholding will simply eliminate many low-confident predictions on unlabeled images which may include some difficult but informative object examples. This accordingly prevents the encoded knowledge to be distilled to the student. To solve this problem, we employ a customized detection loss, \ie the focal loss \cite{lin2017focal} to formulate the consistency regularization between teacher and student predictions, which preserves useful information from unlabeled images as much as possible. 

Specifically, the data imbalance in SSOD includes the following two aspects: the well-matching object detections between the teacher and student, as well as easy background predictions. The large quantity of such examples dominants the consistency loss, which suppresses the contribution of informative training examples. The focal loss is devised to reward the hard examples but penalizes easy ones. So, in our case, it can not only alleviate the negative effects from the large number of easy object and background predictions, but also consolidate the difficult examples, \ie the poor-matching predictions between teacher and student.

We have evaluated the performance of our TSE-T model on two standard large scale benchmarks VOC and COCO. Both evaluation results have shown that TSE-T model can obtain remarkable improvements compared to its lower-bound, the fully-supervised detection model only using labeled images. Specifically, the mAP of our method achieves $80.73\%$ and $40.52\%$ on VOC2007 test set and COCO2014 \textit{minival5k} set respectively, outperforming the baseline by $2.37\%$ and $1.49\%$. It should noted that our method sets the state-of-the-art performance in SSOD on VOC2007 benchmark.

We summarize our contributions as follows:

\begin{itemize}
\item[1] We formally employ the KD framework in SSOD task which constructs a well-trained teacher to regularize the training of a student using unlabeled images.
\item[2] We propose TSE-T model which simultaneously ensembles the data and model diversities. This method produces better targets to train the student but does not significantly increase computational complexity.
\item[3] We use focal loss to solve the data imbalance problem, which results in an efficient and effective usage of unlabeled images in SSOD.
\end{itemize}

The rest of the paper is organized as follows. We review related works in Section \ref{sec::relatedwork}. We elaborate the proposed method in Section \ref{sec::method}. We describe experimental results in Section \ref{sec::experiments}. Finally, in Section \ref{sec::conclusions} we conclude our work and present several potential directions for future work.

\begin{figure*}[!t]
  \centering 
  \includegraphics[width=\textwidth]{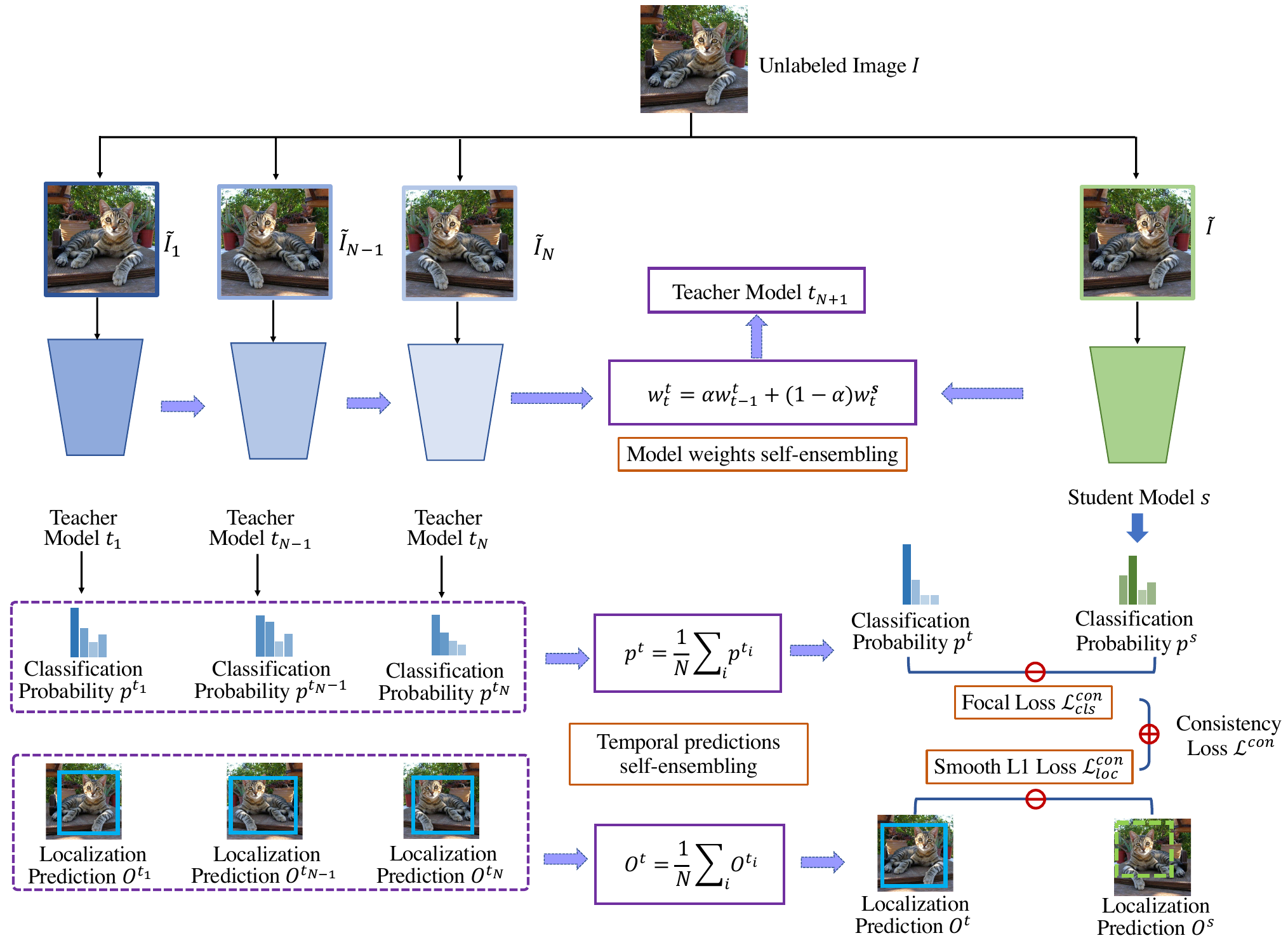}
  \caption{\textbf{A Detailed Graphical Illustration for the Proposed TSE-T Model}. Our method is made on top of the KD framework which consists of a teacher and a student model. Our TSE-T model is devised to ensemble the temporal teacher predictions and ensemble the teacher model weights with the student model weights. These self-ensembling methods yield better targets for unlabeled images which can better retrain the student to improve its performance. We use orange bounding boxes to indicate our main contributions.}
\label{fig::fig2}
\end{figure*}

\section{Related Works}
\label{sec::relatedwork}
In this section, we review the related topics to our work, including object detection (Section \ref{subsec::od}), semi-supervised learning models (Section \ref{subsec::semi}) and model ensemble (Section \ref{subsec:me}).

\subsection{Object detection}
\label{subsec::od}
Object detection is one of the most active research topics in computer vision community \cite{viola2001rapid}. There have been developed hundreds of well-performed detectors. In this work, we focus on the generic object detection models using deep learning \cite{liu2020deep}. The pioneered work R-CNN used the deep learning methods to extract features in the conventional object detection pipeline \cite{girshick2014rich}. The Fast-RCNN \cite{girshick2015fast} and Faster-RCNN \cite{ren2015faster} initiated the study on typical two-stage detectors which successfully implemented the object detection with an end-to-end deep learning architecture. The FPN \cite{lin2017feature} and RetinaNet \cite{lin2017focal} improved the feature representation for object detection by using a decoder-like feature pyramid. To be continued, one-stage detectors, including DenseBox \cite{huang2015densebox} and SSD \cite{liu2016ssd}, were developed, which generate dense predictions using fully convolutional neural networks. This type of methods are much faster, one extraordinary trend of which refers to as the YOLO \cite{redmon2016you,redmon2017yolo9000,redmon2018yolov3}.  The mask-rcnn \cite{he2017mask} proposed the multi-task network integrating the object detection and semantic segmentation which reformatted the instance segmentation. All the above methods are using the popular anchor boxes to encode the object bounding box leading to a translation-invariant detection and relieve the difficulty of regression. Recent developed anchor-free detectors \cite{law2018cornernet,zhou2019bottom,zhu2019feature,duan2019centernet} reformulated the object detection as a key points detection and grouping task. This line of object detection methods reduces the quantity of output but still achieves comparable performance.

\subsection{Semi-supervised learning}
\label{subsec::semi}
The semi-supervised learning (SSL) is one important category of machine learning techniques \cite{zhu2005semi,zhu2009introduction,chapelle2009semi}, which aims to train a machine learning model by using a limited number of labeled data and a large amount of unlabeled data. The key to the semi-supervised learning models is to obtain a better prediction on the unlabeled data. Since the emergence of knowledge distillation network \cite{hinton2015distilling}, the semi-supervised learning has been reshaped based on the teacher-student model architecture. An well-posed prediction for the unlabeled data becomes possible using a cumbersome teacher model, and the result is used to guide the training of a light-weighted teacher model. The $\Gamma$ model devised one clean branch and one noisy branch, which learned an auxiliary mapping between the two branches for denoising \cite{rasmus2015semi}. The $\Pi$ model tried to stabilize the predictions obtained from stochastic data transformation and network perturbations. The subjective was to minimize the predictions difference of the same data when introducing various stochastic transformations and passing the data through perturbed networks. The temporal ensembling model improved the prediction for the unlabeled data by accumulating the predictions during training \cite{laine2016temporal}. The ensembling of multiple networks have been proved to be an effective strategy to produce more accurate predictions \cite{russakovsky2015imagenet}. In the field of SSL, the temporal self-ensembling during training may provide with better predictions for the unlabeled images which can be used better targets to train the student. Instead of ensembling the predictions, the mean teacher model \cite{tarvainen2017mean} ensembled the temporal teacher model weights and the student model weights to yield a dynamic teacher model that can learn from the student. This resulted in a temporally evolved teacher model, so the predictions of unlabeled images from the teacher and student model became diverse which is advantageous to the training of the student.

\subsection{Semi-supervised object detection}
A successful trial on semi-supervised object detection using deep learning techniques was the CSD model which adapted the $\Pi$ model to construct the consistent regularization for the detection of the unlabeled image and its augmentation. The CSD was evaluated on VOC dataset and achieved the state-of-the-art performance. A very recent semi-supervised method developed a proposal-based learning scheme for two-stage object detectors \cite{tang2020proposal}. For the original data and its noisy counterpart, the method used a self-supervised proposal learning module to learn consistent perceptual semantics in feature space and consistent predictions. The method was only evaluated on COCO dataset and has achieved similar results compared to the omni-supervised object detection \cite{radosavovic2018data}. Our work bears a certain resemblance to the omni-supervised object detection. This work used two-stage detectors as detection model and proposed a bounding box voting strategy to generate the a hard-label teacher prediction.  Compared to this work, our method is prioritized in the following aspects. (1) Our method keeps the teacher model dynamic to learn from the student by ensembling its temporal model weights and the student model weights. Such model weights ensembling method ensures the diversity of the teacher model, together with the predictions ensembling improving the predictions for unlabeled images. (2) Instead of using hard-label as target to train the student, our method uses soft-label to retain the information from the unlabeled images as much as possible, which is more informative and efficient to train the student \cite{phuong2019towards}. (3) We use focal loss to solve the data imbalance problem caused by dense predictions in SSOD.

\subsection{Model ensembling}
\label{subsec:me}
Model ensembling is an efficient method to improve the performance of a machine learning system because different model holds distinct generalizbility for the same data and an ensemble of multiple models jointly enhance the generalization ability of the whole system. Such methods are widely used in various computer vision applications, for example, in large scale image recognition \cite{russakovsky2015imagenet,he2016deep,hu2018squeeze,beluch2018power}. Model ensembling often employs multiple models that are either trained with different initialization state or configured with different architectures. As for the SSL which uses a teacher-student framework, a drawback of applying the multiple models ensembling is that the computation complexity increases dramatically for both training and inference. To address this issue, the temporal self-ensembling models have been studied \cite{reed2014training,laine2016temporal,french2017self}. This type of methods takes advantages of self-ensembling which aggregates the model weights or a sequential predictions from the latest training epochs. The involvement of a single model during training naturally reduces the computation complexity and model size compared to the previous ensemble methods.

\section{Methodology}
\label{sec::method}

In this section, we firstly present our SSOD problem setup in Section \ref{subsec::problem}, then elaborate the reasons for the selection of baseline detector in Section \ref{subsec::network}, and finally present in detail our proposed TSE-T approach in Section \ref{subsec::model}.

\subsection{SSOD Problem Setup}
\label{subsec::problem}
An overall framework of our proposed SSOD system is illustrated in Fig. \ref{fig::fig1}. The objective of our proposed SSOD approach is to distil knowledge from geometrically transformed unlabeled images without the requirement of training a large set of models. Our pipeline (Fig. \ref{fig::fig1}) involves the following steps:
\begin{enumerate}
  \item pretrain an object detector on labeled dataset in a fully-supervised manner, and use it to initialize the teacher and the student;
  \item apply the teacher model to a number of geometric transformations of unlabeled samples to generate detections for the unlabeled samples;
  \item ensemble multiple teacher predictions on the unlabeled data to automatically generate training targets (soft labels) for student;
  \item retrain the student on the union of the manually labeled data and automatically labeled data;
  \item update the teacher by ensembling its temporal model weights and current student model weights.
\end{enumerate}

Specifically, our method is based on the KD framework,  consisting of two models: a teacher and a student, both of which are initiated from a typical one-stage detector such as the RetinaNet \cite{lin2017focal} and are pretrained using a certain amount of labeled images in a fully supervised fashion. At semi-supervised training time, the teacher predictions on unlabeled data are used as ``annotations'' to retrain the student in unsupervised manner. It should be noted that the labeled data is also used to train the student in supervised manner to leverage and stabilize the unsupervised training.

The unsupervised retraining of student model using unlabeled data is achieved by a consistency regularization which routinely takes the form of minimizing Mean Squared Error (MSE) between the teacher and student predictions. In this way, the teacher model distils useful knowledge, \ie the object category and localization in unlabeled images, to retrain the student model. In other words, the knowledge encoded by teacher is decoded in such a way that the student back-propagates the gradients to optimize its parameters which can better generalize the unlabeled images. After training, the student model should achieve comparable performance to that of the  teacher, and is deployed to detect objects in unseen images during testing.

In order to obtain better teacher predictions on unlabeled data, we propose T-SET model which is detailed in Fig. \ref{fig::fig2}, including two types of self-ensembling strategies. (1) We run a temporal series of teacher model on different geometric transformations (\eg, flipping) of an unlabeled image, and then ensemble their predictions as the final training targets for the student, because such multiple temporal model ensembling and data augmentation manner takes the advantages that different temporal model is equipped with different generalization ability on different data transformation. When ensembling the diverse models and data, the teacher predictions on unlabeled data will be improved with a large margin from the student predictions. (2) We ensemble the temporal model weights of the teacher with the student model weights using the exponential moving average, so that the teacher can gradually learn from the teacher to not only enhance its performance but increase its temporal diversity.

Now we formally define the SSOD problem setup. Suppose we are given a dataset of $M$ images $\mathcal{D}=\{\mathbf{I}^{(i)}\}_{i=1}^M$. For a labeled image $\mathbf{I} \in \mathcal{D}$, $\textbf{\emph{y}}=[P_x, P_y, P_w, P_h, c]^T$ is the ground truth label vector which defines the category $c \in [1,...,C]$ of an object and specifies the pixel coordinates $(P_x, P_y)$ of the center of object bounding box together with its width and height $(P_w,P_h)$ in pixels (we drop the superscript $i$ unless it is needed). For an unlabeled image $\mathbf{I}^u \in \mathcal{D}$, we use the teacher prediction as the target labels (\ie, pseudo soft labels). For instance, one detected object bounding box $O$ from an unlabeled image is specified in the same way, represented as $\hat{\textbf{\emph{y}}}^t=[O_x, O_y, O_w, O_h, p]^T$, where $p$ is the class probability.

In SSOD, we aim to promote the performance of the student model regularized by the teacher model using the unlabeled images. This is achieved in an unsupervised manner and defined as consistency regularization (unsupervised loss) between the teacher and student predictions $ \mathcal{L}^{con}[\hat{\textbf{\emph{y}}}^s, \hat{\textbf{\emph{y}}}^t]$, where $ \hat{\textbf{\emph{y}}}^s$ is the student prediction which is formulated in the same way of the teacher prediction. If only the unlabeled images are used to retrain the student model, it may lead to an ill-posed convergence behavior of the student.  For a balanced semi-supervised training, we also employ the labeled images in the form of a supervised loss $\mathcal{L}^{sup} [\hat{\textbf{\emph{y}}}^s, \textbf{\emph{y}}]$. We will specify the definition of $\mathcal{L}^{con}$ and $\mathcal{L}^{sup}$ in following sections. The objective of the SSOD is to optimize the student model to minimize both the unsupervised loss and supervised loss $\mathcal{L}=\mathcal{L}^{sup} [\hat{\textbf{\emph{y}}}^s, \textbf{\emph{y}}] + \mu_1 \mathcal{L}^{con} [\hat{\textbf{\emph{y}}}^s, \hat{\textbf{\emph{y}}}^t]$, where the hyper-parameter $\mu_1$ will be specified later.

\subsection{Baseline Detector}
\label{subsec::network}

Object detectors can be generally classified into two categories: one-stage  \cite{lin2017focal,liu2016ssd,redmon2016you} and two-stage  \cite{ren2015faster,dai2016r,he2017mask}. The main distinction is that the two-stage detectors employ a region proposal network (RPN) to explicitly generate object candidates. Next the non-maximum suppression (NMS) merges the spatially duplicated prediction candidates with a certain amount of overlap. In this work, we choose to use one-stage detector, for example, the RetinaNet \cite{lin2017focal}, for the following concerns.

On one hand, the SSOD attempts to preserve the predictions as many as possible for each default location. The underlying intuition is that the objective of our TSE-T is to synchronize the predictions of the student with those of the teacher, so that the student can approach the performance of a well-trained teacher. So, we do not hope to employ NMS before the fine-grained object detection both in teacher and student, because many confident predictions may be suppressed and the knowledge distillation from the teacher may deteriorate the generalizability of the student model for those predictions. 

On the other hand, in two-stage detectors, it is relatively difficult to solve the matching problem in RPN between an image and its transformations. This is because the employment of NMS in RPN results in the misalignment of the region proposals for different input images. Simple solution tackles this problem by only feeding the original image into RPN and use the location of the obtained region proposals to estimate the location of region proposals from the transformed image \cite{jeong2019consistency}. It has empirically found that the adaptation of two-stage detectors in SSOD performs worse than one-stage detectors due to the lack of consistency regularization in RPN training \cite{jeong2019consistency}. However, we would like to retain the potentials of our method in two-stage detectors if a proper solution can be found to solve the matching problem.

\subsection{TSE-T Model}
\label{subsec::model}

\subsubsection{Ensemble temporal predictions}
\label{subsubsec::predictions}

A well-performed teacher model in SSOD should provide with better predictions of objects presented in unlabeled images. These predictions should remain sufficient dissimilarities from the student predictions, such that the knowledge encoded in these potential objects can be fully captured and then distilled to the student. By pushing the student to obtain consistent predictions as the teacher, the student can improve itself on generalizing the latent objects in unlabeled images. In our proposed TSE-T, we achieve this purpose by ensembling the temporal teacher predictions from the latest training epochs. Because we add random perturbations for each image in each epoch, this self-ensembling produces a large number of data combination for teacher models at different checkpoint, and this data and temporal model diversity ensures better teacher predictions on unlabeled images.

Specifically, at training time, given an unlabeled image $\mathbf{I}^{\mathbf{u}} \in \mathcal{D}^{\mathbf{u}}$, we retrieve its previous teacher predictions from the last $N$ epochs $\hat{\textbf{\emph{y}}}^{t_1}, \cdots,\hat{\textbf{\emph{y}}}^{t_N}$. The TSE-T model obtains the current teacher prediction by averaging these predictions:

\begin{equation}
\label{eq::teacher}
\hat{\textbf{\emph{y}}}^t=\frac{1}{N}\sum_{i=1}^N \hat{\textbf{\emph{y}}}^{t_i},
\end{equation}

which can be separately denoted as ensembling of the localization and classification.

\begin{equation}
\label{eq::ensembleing1}
\left\{
\begin{aligned}
&O^{t}=\frac{1}{N}\sum_{i=1}^N O^{t_i}, \\[3pt]
&p^t=\frac{1}{N}\sum_{i=1}^N p^{t_i}.
\end{aligned}
\right.
\end{equation}

Due to the employment of the data augmentation, we need to align the predictions before ensembling. We implement this by tracing the image orientation during augmentation and flip the predictions back to the original reference image. We assume that ensembling more teacher predictions from different training epochs may generate better training targets for student. We will validate this assumption and show the effect of varying value of $N$ for SSOD in Section \ref{sec::experiments}.

\subsubsection{Ensemble model weights}
\label{subsubsec::ema}

In the original temporal ensembling model \cite{french2017self}, the teacher model serves as the student model as well. When given an unlabeled image and its transformations, the teacher and student predictions may be very close, for example, if the objects in the image are easy to recognize. These similar predictions contribute little to the consistency regularization which constraints the upper-bound of the student model. Therefore, we expect to decouple the teacher model from the student during semi-supervised training and keep the teacher model evolved instead of fixed as proposed from the original KD framework. To this end, our TSE-T model proposes a temporally updated teacher model which is devised to ensemble the historical teacher model weights with the current student model weights using a momentum term formulated as follows:

\begin{equation}
\label{eq::ema}
w^t_{t}=\alpha w^t_{t-1} + (1-\alpha) w^s_{t},
\end{equation}

where $w^t_{t}$ and $w^t_{t-1}$ denote teacher model weights at current and previous training step respectively; $w^s_t$ denotes the student model weights updated at current training step. $\alpha$ is a momentum parameter to leverage the contribution of previous teacher model weights and current student model weights in updating the current teacher model. Such model weights ensembling method is also referred to as the exponential moving average (EMA). This self-ensembling can be seen as an imitation of a real education circumstance, in which teacher may bias in existing knowledge due to its empiricism or may miss to learn new updated knowledge. In this case, student may help the teacher to escape from this trap by directly conveying its knowledge to the teacher. We note that our teacher model is already more advantageous than the teacher in previous methods, so a slightly additive knowledge transfer from the student may result in a fast convergence of the teacher. This can be manipulated by setting a relatively large value of parameter $\alpha$, for example $\alpha=0.99$ \cite{he2019momentum}.

\subsubsection{Loss functions}
\label{subsubsec::loss}

The total loss of the proposed SSOD framework is defined as follows:

\begin{equation}
\label{eq::loss}
\mathcal{L}=\mathcal{L}^{sup} \left[ \hat{\textbf{\emph{y}}}^s, \textbf{\emph{y}} \right] + \mu_1 \mathcal{L}^{con} \left[ \hat{\textbf{\emph{y}}}^s, \hat{\textbf{\emph{y}}}^t \right].
\end{equation}

We use the hyper-parameter $\mu_1$ to leverage the contribution of the supervised loss $\mathcal{L}^{sup}$ and unsupervised loss $\mathcal{L}^{con}$. The selection of $\mu_1$ will be discussed later in Section \ref{sec::experiments}.

\textbf{Detection loss:} The training objective of object detection is to minimize the prediction errors both for classification and localization. So, we specify the above two objectives as:

\begin{equation}
\label{eq::detection_loss}
\left\{
\begin{aligned}
& \mathcal{L}^{sup} \left[ \hat{\textbf{\emph{y}}}^s, \textbf{\emph{y}} \right] &=\frac{1}{M_1} \sum \left( \mathcal{L}_{cls}^{sup}+\mu_2 \mathcal{L}_{loc}^{sup} \right) \\[3pt]
&\mathcal{L}^{con} \left[ \hat{\textbf{\emph{y}}}^s, \hat{\textbf{\emph{y}}}^t \right] &=\frac{1}{M_2} \sum \left(  \mathcal{L}_{cls}^{con}+\mu_2 \mathcal{L}_{loc}^{con}  \right),
\end{aligned}
\right.
\end{equation}

where $\mathcal{L}^{sup}$ and $\mathcal{L}^{con}$ denote the supervised loss for the labeled images and consistency loss \footnote{Here we use the terms ``consistency loss'', ``unsupervised loss'', ``consistency regularization'' to indicate the identical meaning that defines the consistency measure between the teacher and student predictions on unlabeled images. } for the unlabeled images respectively. $M_1$ and $M_2$ are respectively the total predictions of labeled and unlabeled images. The hyper-parameter $\mu_2$ balances the contribution of the classification and localization loss to the total loss, which will be discussed in Section \ref{sec::experiments}.

In our method, we use focal loss to address the data imbalance problem during training. The employment of focal loss has another advantage that aligns the definition of the supervised and unsupervised loss, separately formulated as:

\begin{equation}
\label{eq::focal}
\left\{
\begin{aligned}
& \mathcal{L}_{cls}^{sup}=-(1-p^s)^\gamma \log(p^s) \\[3pt]
& \mathcal{L}_{cls}^{con}=-(|p^t-p^s|)^\gamma p^t \log(p^s).
\end{aligned}
\right.
\end{equation}

The loss functions retain the form of the standard cross entropy loss, where $p^t$ and $p^s$ are the teacher and student prediction probability respectively for the object class.

As for the localization, we introduce the Smooth L1 loss both for the supervised localization loss and consistency localization loss.

\begin{equation}
\label{eq::localization}
\left\{
\begin{aligned}
& \mathcal{L}_{loc}^{sup}=smooth_{L1} (\tilde{O}^s-\tilde{P}) \\[3pt]
& \mathcal{L}_{loc}^{con}=smooth_{L1} (\tilde{O}^s-\tilde{O}^t),
\end{aligned}
\right.
\end{equation}

where $\tilde{P}$, $\tilde{O}^s$ and $\tilde{O}^t$ are the offsets from the ground-truth, student prediction and teacher prediction to the anchor boxes respectively. We provide an example for the computation of the offsets using the teacher prediction $O^t=[O_x,O_y,O_w,O_h]$.

\begin{equation}
\label{eq::normalize}
\left\{
\begin{aligned}
&  \tilde{O}_x = \left(O_x-d_x \right) / d_w  \\[3pt]
&  \tilde{O}_y = \left(O_y-d_y \right)  / d_h \\[3pt]
&  \tilde{O}_w = \log \left( O_w/d_w \right) \\[3pt]
&  \tilde{O}_h= \log \left( O_h/d_h \right),
\end{aligned}
\right.
\end{equation}

where $d=[d_x, d_y, d_w,d_h]$ is the localization of one anchor box, and $\tilde{O}^t=[\tilde{O}_x, \tilde{O}_y,\tilde{O}_w,\tilde{O}_h]$ is the normalized teacher prediction of one object's localization.

In our settings of SSOD, we use one-stage detection network and avoid employing the NMS before model ensembling. For the unlabeled images, this encourages the emergence of a large number pairwise teacher-student predictions which remain confident consistency. It should be noted that the accumulation of the well-fitted predictions is probable to suppress the inconsistent prediction pairs that are minority in training examples but should be the main contributors in the loss. Thus we revise the the Smooth L1 function to alleviate this effect, which is formulated as follows. 

\begin{equation}
\label{eq::smooth_l1}
smooth_{L1}(x)=
\left \{
\begin{aligned}
& \frac{|x|^3}{3}, & |x|<\beta  \\[3pt]
&  |x|-\beta+\frac{\beta^3}{3}, & |x|\geq \beta
\end{aligned}
\right.
\end{equation}

In Algorithm \ref{alg::alg1}, we summarize the whole training procedure of our method in the form of pseudo code. Here we omit the fully-supervised pre-training step using labeled images, so we directly start from a convergent detection model and train the student model using the proposed TSE-T model.

\begin{algorithm} [t] 
\begin{spacing}{1.2}
\caption{Pseudocode of TSE-T model}  
\label{alg::alg1}  
\SetLine 
$f^{s*}$ = TSE-T($f^s$, $f^t$, $w^s$, $w^t$)

\KwIn{Training dataset $\mathcal{D}$  \\ Pre-trained student model$f^s$ \\ Pre-trained teacher model $f_t$ \\ Student model weights $w^s$  \\ Teacher model weights $w^t$}
\KwOut{Optimized student model $f^{s*}$}
\KwInitial{Epochs=$K$}
\For{$k \leftarrow 0$ \KwTo $K-1$}
{ 	
	 \ForEach{Mini-batch $\mathcal{D}_t$}
 	{
 		$\tilde{\mathcal{D}}_t=stochastic-transformation(\mathcal{D}_t)$
 		
		$\hat{\textbf{\emph{y}}}^s_t=f^s(\tilde{\mathcal{D}}_t)$
		
		\If{$\tilde{\mathcal{D}}_t$ is unlabeled}
		{
			\footnotesize{Align teacher predictions in previous N epochs}
			
			$[\hat{\textbf{\emph{y}}}^{t_1}, \cdots, \hat{\textbf{\emph{y}}}^{t_N}]=f_A([\hat{\textbf{\emph{y}}}^{t_1}, \cdots, \hat{\textbf{\emph{y}}}^{t_N}])$
			
			\footnotesize{Ensemble temporal teacher predictions by Eq. \ref{eq::teacher}}
			
			$\hat{\textbf{\emph{y}}}^t=\frac{1}{N}\sum_{i=1}^N \hat{\textbf{\emph{y}}}^{t_i}$
		}
		
		\small{Compute total loss by Eq. \ref{eq::loss}}
		
		$\mathcal{L}=\mathcal{L}^{sup} \left[ \hat{\textbf{\emph{y}}}^s, \textbf{\emph{y}} \right] + \mu_1 \mathcal{L}^{con} \left[ \hat{\textbf{\emph{y}}}^s, \hat{\textbf{\emph{y}}}^t \right]$ 
		
		\small{Update student model by standard SGD}
		
		 $w_t^s=w_{t-1}^s-\lambda\partial \mathcal{L} / \partial w^s$
		 
		\small{Update teacher model by Eq. \ref{eq::ema}}
		
		$w^t_{t}=\alpha w^t_{t-1} + (1-\alpha) w^s_{t}$
		
		\small{For teacher prediction in next epoch}
		
		$\hat{\textbf{\emph{y}}}_t^{t_{N+1}}=f^t_{w=w^t_{t}}(\tilde{\mathcal{D}}_t)$

	}	
	
	\small{Update teacher predictions for next epoch}
	
	$[\hat{\textbf{\emph{y}}}^{t_1}, \cdots, \hat{\textbf{\emph{y}}}^{t_N}] \leftarrow [\hat{\textbf{\emph{y}}}^{t_2}, \cdots, \hat{\textbf{\emph{y}}}^{t_{N+1}}]$
	
}
\end{spacing}
\end{algorithm}

\section{Expreiments}
\label{sec::experiments}

In this section, we conduct experiments to evaluate the performance of the proposed TSE-T model for SSOD. We use two standard benchmarks for object bounding box localization, the VOC \cite{everingham2010pascal} and COCO \cite{lin2014microsoft}. As for the competing methods, we use the fully-supervised RetinaNet as a strong baseline method.  We also use the state-of-the-art method, the CSD \cite{jeong2019consistency}, for a challenging comparison to show the merit of our method under the semi-supervised setup. We implement our method based on the Mask-RCNN benchmark \cite{pytorch_faster}. For a fair comparison, we train, validate and test the RetinaNet using the implementations from the same Mask-RCNN benchmark as well.

\subsection{Configurations}
\textbf{Datasets} We use VOC and COCO datasets in our experiments. For VOC benchmark, we choose to use the VOC2007 and VOC2012, both of which consist of 20 annotated semantic object classes. Following the configuration in \cite{jeong2019consistency}, we fix VOC2007 test set to evaluate and compare our method with state-of-the-art methods. For COCO benchmark, we choose COCO2014 which includes 80 semantic classes and we follow the standard experimental protocol \cite{bell2016inside,lin2017feature,lin2017focal} which uses the COCO \textit{trainval35k} split and uses the \textit{minval5k} split as test set. In VOC and COCO datasets, there are separately two subsets that are not provided with ground-truth annotations, \ie the VOC2012 test set and the COCO unlabeled set. So, we use these two subsets as extra unlabeled images. In Table \ref{tab::tab1} we show detailed information of the datasets. 

\textbf{Experimental setup} We conduct all the experiments using 4 NVIDIA 1080 Ti GPU cards. We use standard SGD optimizer and set the batch size as 8. For the backbone network of RetinaNet, we choose to use ResNet-50 \cite{he2016deep} for the experiments on VOC dataset, and we will validate the performance of ResNe50 and ResNet-101 for the experiments on COCO dataset. Through all the experiments, we use the standard metric of mean average precision (mAP) to evaluate the performance of a method. 

When pretraining the detection model using labeled images, we use 15 epochs and initialize learning rate as 0.005 which is divided by 10 at epoch 5 and epoch 8 separately. When training the student model using unlabeled images, we use 13 epochs and initialize the learning rate as 0.0005 which is divided by 10 at epoch 10. It has been found that for an SSOD system, once the model converges at a local minimum, it will be difficult to reach a global solution in the following training steps. So we carefully design the update strategy for $\mu_1$. In this work, we aim at a stable transition from full-supervised training to semi-supervised training by slowly increasing the weights of the unlabeled data. We thus gradually increase $\mu_1$ from 0.02 to 1.6 and from 0.01 to 0.08 for ResNet-50 and ResNet-101 backbone networks respectively. As for $\mu_2$, we choose the value of 0.07 and 0.1 separately for ResNet-50 and ResNet-101 backbone networks, which modulates the classification and localization loss at a similar scale. Finally, we set $\beta=0.4$ in Eq.\ref{eq::smooth_l1} by validation.

\begin{table}[!t]
\footnotesize
\renewcommand{\arraystretch}{1.3}
\caption{Datasets Statistics}
\label{tab::tab1}
\centering
\begin{threeparttable}
\begin{tabular}{c|ccccc}
\hline
\diagbox [width=6em,trim=l] {\bf{Dataset}}{\bf{Fold}} & \textbf{Train} &  \textbf{Val} & \bf{Train/Va}l & \textbf{Test} &  \textbf{Unlabeled} \\ \hline
\footnotesize VOC2007  & 2,501   & 2,510    & 5011       & 4,952\tnote{*}   & -- \\
\footnotesize VOC2012  & 5,717    & 5,823   & 10,540   & \textit{10,991}\tnote{**}  & -- \\ \hline
\footnotesize COCO       & 80,000 & 35,000  & 115,000 &  5,000 \tnote{*}  & \textit{123,403}\tnote{**}  \\
\hline
\end{tabular}
 \begin{tablenotes}
        \footnotesize
        \item[*] ~Test set in our experiments
        \item[**] Extra unlabeled images in our experiments
      \end{tablenotes}
\end{threeparttable}
\end{table}

\begin{figure}[!t]
  \centering 
  \includegraphics[width=0.4\textwidth]{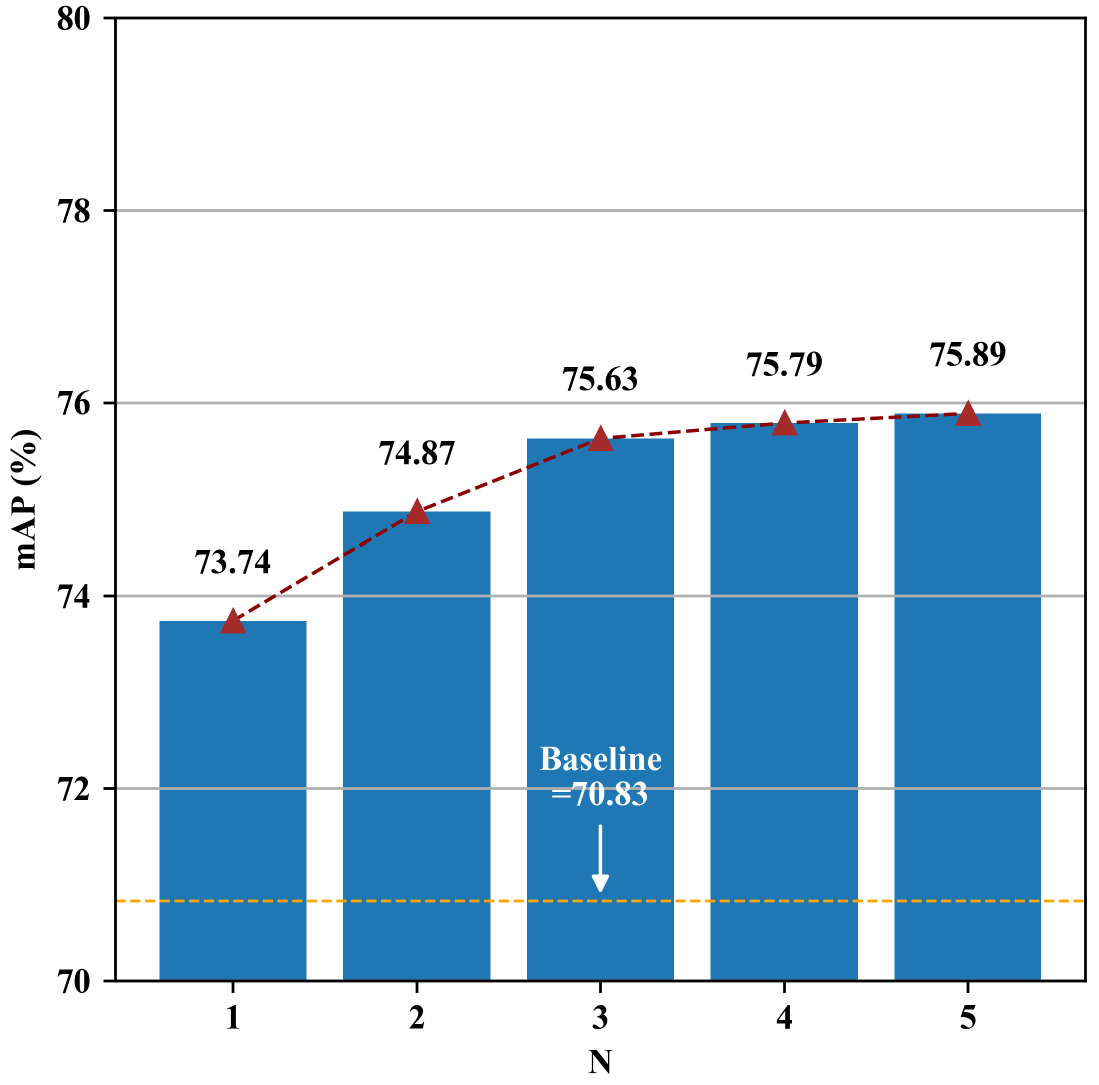}
  \caption{\textbf{Validation of $N$ in TSE-T.} X-axis indicates an $N$ ranging from 1 to 5. Y-axis indicates the mAP of detection results. We use red plotting to show the performance tendency according to $N$. The horizontal dash line denotes the baseline performance of the fully-supervised detector. }
\label{fig::fig3}
\end{figure}

\begin{table*}[!t]
\footnotesize
\renewcommand{\arraystretch}{1.3}
\caption{Detection results on VOC2007 test set}
\label{tab::tab2}
\centering
\begin{threeparttable}
\begin{tabular}{l | M{1.2cm}M{1.2cm}M{1.2cm} | M{1.8cm}M{1.8cm}M{1.8cm} | l } 
\hline
\bf Model & \bf 07train/val & \bf 12train/val & \bf 12test &  \bf Ensemble &  \bf EMA &  \bf Detection Loss &  \bf mAP\\ 
\hline
RetinaNet\tnote & \checkmark & -- & -- & -- & -- & -- & 71.56 \\
TSE-T & \checkmark & SS & -- & -- & \checkmark & -- & 72.45$_{\uparrow 0.89}$\\
TSE-T & \checkmark & SS & -- &  -- & \checkmark  & \checkmark & 74.58$_{\uparrow 3.02}$ \\
TSE-T & \checkmark & SS & -- & \checkmark & --  & -- & 75.11$_{\uparrow 3.55}$\\
TSE-T & \checkmark & SS & -- &  \checkmark  & -- & \checkmark & 75.46$_{\uparrow 3.90}$ \\
TSE-T	 &\checkmark	 & SS	& -- & \checkmark & \checkmark & -- &	76.24$_{\uparrow 4.68}$\\ 
TSE-T	 &\checkmark	 & SS	& -- & \checkmark & \checkmark &  \checkmark &	76.68$_{\uparrow 5.12}$\\ \hline
TSE-T & \checkmark &SS & SS & -- & \checkmark & -- & 73.14$_{\uparrow 1.85}$ \\
TSE-T & \checkmark & SS & SS &-- &  \checkmark &  \checkmark &  75.35$_{\uparrow 3.77}$ \\
TSE-T & \checkmark & SS & SS & \checkmark & --  & -- & 76.05$_{\uparrow 4.49}$\\
TSE-T & \checkmark & SS & SS & \checkmark & --  &  \checkmark & 76.35$_{\uparrow 4.77}$ \\
TSE-T &\checkmark &SS	& SS & \checkmark & \checkmark & -- & 76.98$_{\uparrow 5.42}$ \\ 
TSE-T &\checkmark &SS	& SS & \checkmark & \checkmark & \checkmark & 77.24$_{\uparrow 5.68}$ \\ \hline
RetinaNet & \checkmark & \checkmark & -- & -- & -- & -- & 78.36 \\
TSE-T & \checkmark & \checkmark & SS & -- & \checkmark & -- & 78.87$_{\uparrow 0.51}$ \\
TSE-T & \checkmark & \checkmark & SS & -- & \checkmark  & \checkmark & 78.86$_{\uparrow 0.50}$   \\
TSE-T & \checkmark & \checkmark & SS & \checkmark & --  & -- & 79.37$_{\uparrow 1.01}$\\
TSE-T & \checkmark & \checkmark & SS & \checkmark & --  & \checkmark &  79.76$_{\uparrow 1.40}$  \\
TSE-T &\checkmark & \checkmark & SS & \checkmark & \checkmark & -- & 80.35$_{\uparrow 1.99}$\\
TSE-T &\checkmark & \checkmark & SS & \checkmark & \checkmark & \checkmark  & \bf{80.73}$_{\uparrow 2.37}$\\ \hline
\end{tabular}
\end{threeparttable}
\end{table*}

\subsection{Experiments on VOC dataset}
\label{subsec::ablation}
\textbf{Validation of N.} In this experiment, we validate the effects of $N$ in our proposed TSE-T for SSOD. $N$ determines the number of historical teacher models which are used to ensemble the current teacher predictions on unlabeled images. Here, we use VOC2007 train and validation sets as labeled data and VOC2012 train set as unlabeled data. We leave out VOC2012 validation set to evaluate the performance of our method. We design this setup for a balance between labeled and unlabeled images. Constrained to the computation capacity of our firmware, we set $N$ to range from 1 to 5. For the baseline method, we use the ReinaNet which is trained using the same labeled data, \ie VOC2007 train and validation sets. As for our method, we keep all the configurations stable, for example, the learning rate and training epochs, when $N$ is assigned with different values.

From the validation results shown in Fig.\ref{fig::fig3}, we can see that our method outperforms the baseline method by a large margin. And an increasing $N$ corresponds to a continuous performance improvement, which means ensembling more teacher predictions benefits the re-training of the student model. We notice that a large value of $N$, for example $N=5$, may not sufficiently increase the mAP of TSE-T, but we still choose to use $N=5$ for remaining experiments because it guarantees the best validation performance of our TSE-T model.

\textbf{Ablation study.} In this experiment, we validate the effectiveness of the basic modules in our TSE-T model, \ie (a) the temporal teacher predictions self-ensembling which we denote as ``Ensemble'', (b) the temporal teacher model weights ensembling which we denote as ``EMA'', and (c) the customized detection loss based consistency regularization which we denote as ``Detection loss''. For comparisons, (a) when omits the ``EMA'', we froze the teacher model during semi-supervised training and ensemble its predictions on unlabeled images from $N$ latest training epochs as targets to train the student. (b) When omit the ``Ensemble'' , we only use the teacher predictions on unlabeled images from the latest training epoch as targets to train the student. (c) When omit the ``Detection loss'', we use the standard Euclidean distance to formulate consistency regularization. 

To enable the semi-supervised training using both labeled and unlabeled images, we set the following different configurations for the dataset: (a) 2007train/val as labeled images and 2012train/val as unlabeled images; (b) 2007tain/val as labeled images and 2012train/val/test as unlabeled images; (c) 2007train/val+2012train/val as labeled images and 2012test as unlabeled images. Compared to (a), setup in (b) includes more unlabeled images, and setup in (c) includes more labeled images, which respectively represents different training conditions for the SSOD. 

We show the results on VOC2007 test set in Table \ref{tab::tab2}. We use the abbreviation ``SS'' to show that the data is used as unlabeled images in semi-supervised training. For each training setup, we show the baseline performance of the supervised object detector, the RetinaNet, in the first row. From the results, we obtain the following observations.

(1) Comparing the results of our TSE-T model to the baseline method, the fully-supervised object detector RetinaNet, it clearly shows that the semi-supervised training is feasible and effective to improve the performance of an object detector. In particular, our method gains large-margin performance improvement for the object detector trained with a limited quantity of labeled data. For example, when only using VOC2007 train/val as labeled data and using VOC2012 train/val as unlabeled data in training setup (a), the mAP of our method achieves $76.68\%$ which outperforms the baseline performance $71.56\%$ by $5.12\%$.

(2) We can see that each of the basic modules in our TES-T independently improves the performance of SSOD under various training conditions. The concurrency of these basic modules results in the best performance of the detection model. This means that the performance of our TSE-T model is not limited by the upper-bound performance of each basic module; Instead, the intrinsic integration of the proposed strategies cooperatively leads to the dramatic improvement of our method. 

(3) As for the self-ensembling strategies, the temporal teacher predictions ensembling seems gain more performance improvement than the temporal teacher model weights ensembling across all the training setups. For example, in training setup (a), the mAP of solely employing the former self-ensembling strategy achieves $75.11\%$ which exceeds the performance of solely employing the latter self-ensembling strategy $72.45\%$ by $3.01\%$. This result shows that ensembling the temporal teacher predictions significantly improves the teacher predictions on unlabeled images, which ensures a certain gap to the student predictions and accordingly benefits the training of the student. The observation that solely employing the temporal teacher model weights ensembling gains a limited amount of performance improvement for SSOD implies that the teacher may learn limited knowledge from the student.

(4) Comparing the results obtained from training setup (a) and (b), when using more unlabeled images, the performance of our TSE-T model is further improved, the mAP increasing from $76.68\%$ to $77.24\%$. Comparing the results obtained from training setup (b) and (c), when using more labeled images, our TSE-T model gains a large-margin improvement from $77.24\%$ to $80.73\%$. These results suggest that solely increasing the quantity of unlabeled images for an SSOD system may lead the performance improvement to reach a local maximum. Under this situation, the employment of a certain amount of labeled images will guide the detector to escape from this dilemma. The key factor behind is that the supervised training using the extra labeled images lifts the lower-bound of our TSE-T model. One may notice a limitation of our method that when an object detector is better optimized using more labeled data, its performance improvement using our semi-supervised training may be confined within a limited range. However, our proposed TSE-T model still holds its potentials to largely improve the performance of object detection when only a very limited number of labeled images are available.

\textbf{Comparison on VOC2007 test set.} In this experiment, we compare the performance of our TSE-T model with competing baseline methods. We use two types of baseline methods: a strong fully-supervised object detector the RetinaNet \cite{lin2017focal}, and a state-of-the-art SSOD method on VOC2007 test set, the CSD model \cite{jeong2019consistency}. 

In Table \ref{tab::tab3}, we show the experimental results. Because the CSD model uses three types of detectors, SSD-300, SSD-512 \cite{liu2016ssd} and RFCN \cite{dai2016r}, so we show both the original results of these baseline detectors and the results of the CSD model trained on top of these detectors. In Table \ref{tab::tab3}, we show the results of our TSE-T model trained under the same conditions of the CSD model. From the results, we find the following important clues.

(1) The employment of unlabeled images for both SSOD methods , \ie our TSE-T model and the CSD model, indeed improves the performance of the object detector, which results in a remarkable absolute performance improvement compared to the fully-supervised detection model trained using the same amount of labeled images. In the table, we use vertical arrows to indicate the absolute performance increase from the SSOD method to the corresponding fully-supervised detector. 

(2) We find that our TSE-T model trained using less unlabeled images already performs better than the best performed CSD model. The TSE-T model trained only using 2012train/val set as unlabeled images achieves the mAP of $76.68\%$, which outperforms the the best performed CSD model CSD-SSD-512 by $0.88\%$, while the latter obtains the mAP of $75.80\%$ trained with the whole VOC2012 set. When using exactly the same training setup, the mAP of our method achieves $77.24\%$ which exceeds the best performed CSD model by $1.44\%$. We show these comparisons using italic digits in brackets from Table \ref{tab::tab3}.

(3) We note that the performance of baseline detectors of our TSE-T model and the CSD model are different. Specifically, the SSD-512 obtains the mAP of $73.30\%$ and the RetinaNet obtains the mAP of $71.56\%$ when using the same training data. This performance difference might affect the performance comparison of our TSE-T model and the CSD model. However, compared to the CSD model, the performance of our TSE-T model gains much more absolute improvement from its baseline object detector. For example, under the same training setup of using 2012train/val/test set as unlabeled images, our TSE-T model outperforms its baseline by $5.68\%$ and the best performed CSD model outperforms its baseline by $2.50\%$. So, this observation convinces us that our method is more advanced because it boosts the performance of an object detector initialized from an ill-posed starting point.

(4) By employing more labeled images, the mAP of our TSE-T achieves $80.73\%$ which exceeds CSD model by $4.93\%$. This is a remarkable performance improvement which sets the new state-of-the-art performance on the VOC2007 test set under the semi-supervised setup.

\begin{table}[!t]
\footnotesize
\renewcommand{\arraystretch}{1.3}
\caption{Performance Comparison on VOC2007 test set}
\label{tab::tab3}
\centering
\begin{threeparttable}
\begin{tabular}{p{1.7cm} M{1.2cm}M{1.2cm}M{0.6cm} l}
\hline
\bf Model & \bf 07train/val & \bf 12train/val & \bf 12test & \bf mAP\\ 
\hline
SSD-300 & \checkmark & -- & -- & 70.20 \\
CSD-SSD-300 &	\checkmark & SS & SS & 72.30$_{\uparrow 2.10}$ \\ \hdashline
RFCN & \checkmark & -- & -- & 73.90 \\
CSD-RFCN &	\checkmark & SS &	SS & 74.70$_{\uparrow 0.80}$ \\  \hdashline
SSD-512 & \checkmark & -- & -- & 73.30 \\
\textit{CSD-SSD-512} &	\checkmark & SS & SS & \textit{75.80}$_{\uparrow 2.50}$\\ \hline
RetinaNet & \checkmark & -- & -- & 71.56 \\
TSE-T	 &\checkmark &  SS & -- &	 76.68$_{\uparrow 5.12(\textit{0.88})}$ \\
TSE-T & \checkmark & SS & SS & 77.24$_{\uparrow 5.68(\textit{1.44})}$\\  \hdashline
RetinaNet & \checkmark & \checkmark & -- & 78.36 \\
TSE-T &\checkmark & \checkmark & SS & \textbf{80.73}$_{\uparrow 2.37(\textit{4.93})}$ \\
\hline
\end{tabular}
% \begin{tablenotes}
%        \footnotesize
%        \item[*]~~The baseline performance to compare CSD model.
%        \item[**]~~The baseline performance to compare our TSE-T model.
%      \end{tablenotes}
\end{threeparttable}
\end{table}

\subsection{Experiments on COCO dataset}

\begin{table*}[!t]
\footnotesize
\renewcommand{\arraystretch}{1.3}
\caption{Performance Evaluation of Varying Backbone Networks on COCO dataset}
\label{tab::tab4}
\centering
\begin{threeparttable}
\begin{tabular}{l l | M{1cm} M{1cm} M{1cm} | l l l l l l}
\hline 
\bf{Model} & \bf{Backbone} & \bf{train} & \bf{val} & \bf{unlabeled} & \bf{AP} & \bf{AP}$_{50}$ & \bf{AP}$_{75}$ &\bf{AP}$_{S}$ & \bf{AP}$_{M}$ & \bf{AP}$_{L}$   \\ 
\hline
RetinaNet &  Resnet50 & \checkmark & -- & -- & 34.51 & 53.26 & 36.54 & 17.96 &	37.29 & 46.56  \\
TSE-T & Resnet50 & \checkmark &	 SS &	-- & 35.42$_{\uparrow 0.91}$ & 53.88$_{\uparrow 0.62}$ & 37.40$_{\uparrow 0.86}$ & 18.87$_{\uparrow 0.91}$ & 40.16$_{\uparrow 2.87}$ & 48.70$_{\uparrow 2.14}$  \\ \hline
RetinaNet &  Resnet50 & \checkmark & \checkmark & --& 36.34 & 55.22 & 38.90 & 19.66 & 39.94 & 48.95  \\
TSE-T & Resnet50 & \checkmark & \checkmark & SS & 36.96$_{\uparrow 0.62}$ & 55.70$_{\uparrow 0.48}$  & 39.42$_{\uparrow 0.52}$  & 19.59$_{\downarrow 0.07}$  & 40.76$_{\uparrow 0.82}$ & 50.12$_{\uparrow 1.17}$  \\
\hline
RetinaNet & Resnet101 & \checkmark & \checkmark & --  & 39.03 &	 58.31 &	41.66 &	22.01 &	42.83 &	51.87 \\
TSE-T & Resnet101 & \checkmark & \checkmark & SS & 40.14$_{\uparrow 1.11}$ &	59.58$_{\uparrow 1.27}$ &	42.78$_{\uparrow 1.12}$ & 23.93$_{\uparrow 1.92}$ &	44.70$_{\uparrow 1.92}$  &	50.99$_{\downarrow 0.88}$  \\ 
%
%\hline
%RetinaNet\tnote{*}     & Resnet101 & \checkmark & \checkmark & --  & 40.06 &	 59.55 &	42.86 &	23.17 &	44.04 &	52.89 \\
%
TSE-T\tnote{*}     & Resnet101 & \checkmark & \checkmark & SS & \bf{40.52}$_{\uparrow 1.49}$ &	\bf{59.93}$_{\uparrow 1.62}$ &	\bf{43.48}$_{\uparrow 1.82}$ & \bf{24.13}$_{\uparrow 2.12}$ &	\bf{45.47}$_{\uparrow 2.64}$  &	\bf{52.97}$_{\uparrow 1.10}$ \\
\hline
\end{tabular}
 \begin{tablenotes}
        \footnotesize
        \item[*]~~Use extra training image augmentation, i.e. random image resizing.
      \end{tablenotes}
\end{threeparttable}
\end{table*}

Considering that the COCO dataset is a more challenging object detection benchmark, we conduct experiments to find out an efficient backbone network for the detection model. Here, we choose to use Resnet50 and Resnet101 for comparison. We use the standard evaluation metrics for COCO dataset to illustrate the results: AP (averaged average precision over varying thresholds of IoU), AP$_{50}$ (AP of IoU=0.5), AP$_{75}$ (AP of IoU=0.75), AP$_S$ (AP for ``small size''  objects), AP$_M$ (AP for ``medium size''  objects), and AP$_L$ (AP for ``large size'' objects). As for the training data, we configure two different setups: (a) train set as labeled images and val set as unlabeled images; (b) train/val set as labeled images and the extra unlabeled images as unlabeled data. In Table \ref{tab::tab4}, we show the experimental results of the fully-supervised object detector and our TSE-T model under various training setups. From the results, we can draw the following conclusions. 

(1) By comparing TSE-T with RetinaNet, we can see that our TSE-T model outperforms its fully-supervised counterpart on COCO benchmark when using the same backbone network in detection model and using the same training data. For example, when using Resnet50 as backbone network, the AP of our TSE-T model achieves $35.42\%$ and $36.96\%$ under training setup (a) and (b) separately, which outperforms the RetinaNet trained under the same conditions by $0.91\%$ and $0.62\%$ respectively. When using Resnet101 as backbone network, out method outperforms the baseline method by $1.11\%$,  which suggests that our TSE-T model is generic to improve the performance of SSOD regardless the specific type of backbone network. 

(2) When using more labeled images to train the RetinaNet on COCO dataset, its performance obtains a remarkable improvement, whose AP achieves $39.03\%$. This accordingly improves the lower-bound performance of our TSE-T model whose AP finally achieves $40.14\%$. To further improve the performance of our method, we use random resizing to augment the training images. The results are shown in the last row in Table \ref{tab::tab4} indicated by a asterisk. In this case, the AP of our TES-T model has achieved $40.52\%$ which exceeds the fully-supervised baseline by $1.49\%$. 

(3) We observe a phenomenon that our TSE-T model with a deeper backbone network like Resnet101 trained using more labeled data can obtain more performance improvement on detecting small and medium size objects. Such behavior of our TSE-T model may imply that the difficult examples in the objects with small and medium size can be properly decoded and distilled to the student under such a training setup. We will further reasoning and generalize this behavior on other training setups of COCO dataset in future work.  

\subsection{Qualitative results}
\begin{figure*}[!t]
  \centering
    \subfloat[Detection results from VOC2007 test set]{\includegraphics[width=\textwidth]{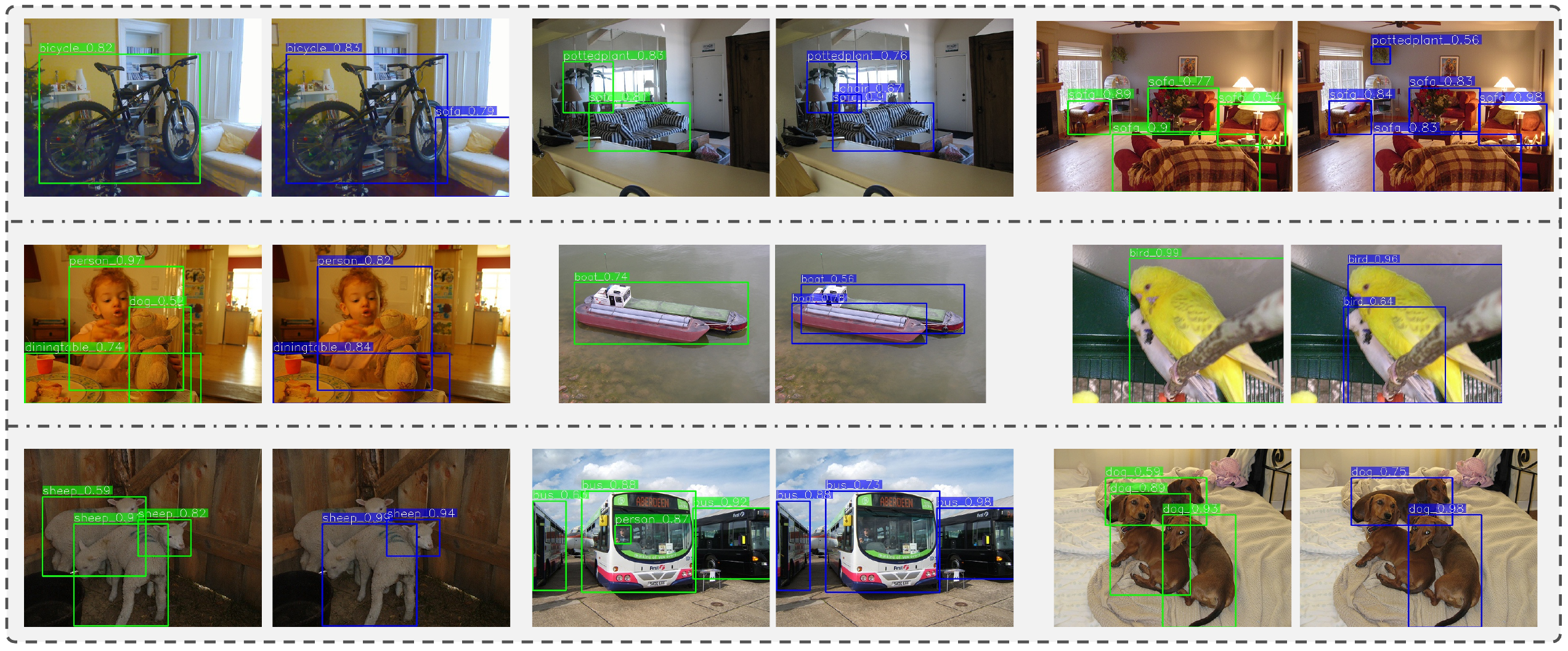}} \\
	\subfloat[Detection results from COCO2012 test-dev set]{\includegraphics[width=\textwidth]{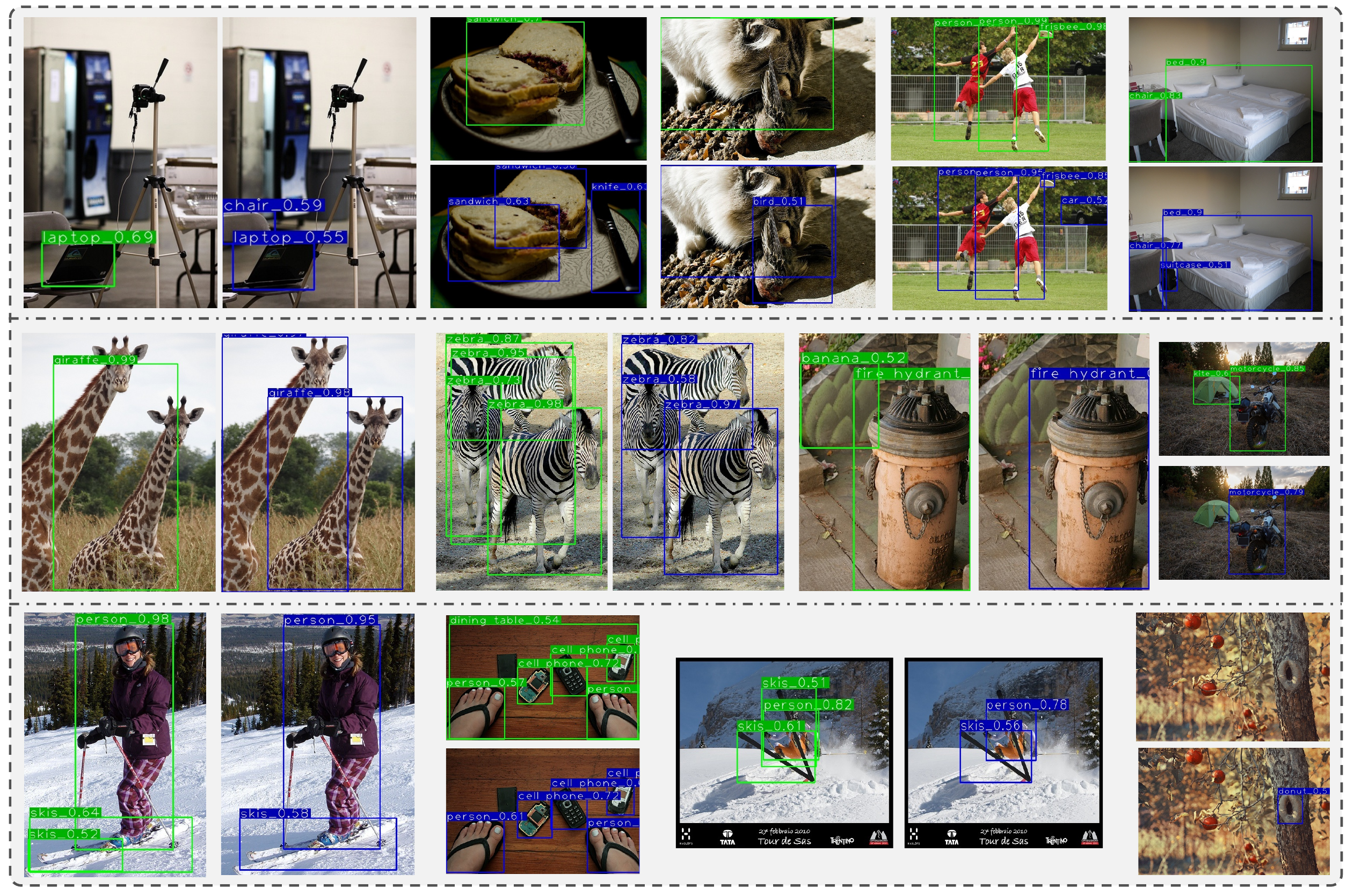}}
  \caption{\textbf{Detection results comparison of TSE-T model and its fully supervised counterpart, the RetinaNet, on VOC and COCO datasets}. The green bounding boxes indicate the detections from the RetinaNet, and the blue bounding boxes denote the detections of our TSE-T model. We arrange the detection results of the same image side by side for a convenient read. For each dataset, we show the the examples from following cases: The TSE-T model recalls difficult objects (Top row); The TSE-T model alleviates false positives (Middle row); The TSE-T may fail to detect the objects with severe occlusions (Bottom row).}
	\vspace{0.4in}
\label{fig::fig4}
\end{figure*}

In Fig. \ref{fig::fig4} (A) and (B), we visualize the detection results from VOC2007 test set and COCO2014 \textit{minval5k} set. We organize the detection results from an image obtained by the RetinaNet and our TSE-T model side by side for an easy comparison. We summarize and show several different cases for a fair and comprehensive comparison. Case I: Our method can successfully detect the small difficult objects. Case II: Our method can alleviate the false positive detections that the RetinaNet misclassified. Case III: Some extreme examples that our method fails to detect. We show these different detection results in top, middle and bottom row of Fig. \ref{fig::fig4} (A) and (B) respectively. These visual effects demonstrate the effectiveness of our TSE-T model to improve object detection under semi-supervised setting. The self-ensembling strategies and the employment of focal loss in our method formulate a better teacher model which yields better predictions on unlabeled images for difficult examples, for example, small objects with severe shape deformations and the objects with occlusions. Among the detection results, we find that our method fails to separate the ski boards from each other and fails to recognize some animals with severe occlusions like the dog and sheep. We assume this phenomenon is caused by the class imbalance in the unlabeled images which may be solved in future work by taking the quantity of training examples from each class into account.

\section{Conclusions}
\label{sec::conclusions}
We propose the TSE-T model to tackle the challenge of SSOD. We have two fundamental goals for TSE-T. First, based on the KD framework, the student is regularized by the teacher to better generalize the latent objects in unlabeled images. Second, the student needs to intimate the whole behavioral patterns of the teacher on predicting the unlabeled images rather than only learning high-confident predictions from the teacher. To these ends, the proposed TSE-T model first ensembles temporal teacher predictions and temporal teacher model weights, which increases data and model diversity. This produces better teacher predictions which hold a large gap to the student predictions and accordingly lifts the upper-bound to optimize the student. Moreover, TSE-T adapts the focal loss to formulate the consistency loss between teacher and student predictions. Such method retains all useful information, such as the information encoded in low-confident hard examples from unlabeled images, which aligns the behaviors of teacher and student and naturally mitigates the class imbalance problem in object detection. Experimental results show that our method sets the new state-of-the-art performance of SSOD on the VOC2007 test set and has obtained a dramatic improvement on COCO2014 \textit{minval5k} set, the mAP of which achieves $80.73\%$ and $40.52\%$ separately. A possible direction to further improve our work may refer to a balance between ensembling multiple heterogeneous models and training efficiency. On the other hand, we could take the categorical balance in unlabeled images into account and apply other augmentations to leverage the detection of objects with large scales.

%\ifCLASSOPTIONcaptionsoff
%  \newpage
%\fi

\bibliographystyle{IEEEtran}
\bibliography{IEEEabrv,reference}

\begin{thebibliography}{10}
\providecommand{\url}[1]{#1}
\csname url@samestyle\endcsname
\providecommand{\newblock}{\relax}
\providecommand{\bibinfo}[2]{#2}
\providecommand{\BIBentrySTDinterwordspacing}{\spaceskip=0pt\relax}
\providecommand{\BIBentryALTinterwordstretchfactor}{4}
\providecommand{\BIBentryALTinterwordspacing}{\spaceskip=\fontdimen2\font plus
\BIBentryALTinterwordstretchfactor\fontdimen3\font minus
  \fontdimen4\font\relax}
\providecommand{\BIBforeignlanguage}[2]{{%
\expandafter\ifx\csname l@#1\endcsname\relax
\typeout{** WARNING: IEEEtran.bst: No hyphenation pattern has been}%
\typeout{** loaded for the language `#1'. Using the pattern for}%
\typeout{** the default language instead.}%
\else
\language=\csname l@#1\endcsname
\fi
#2}}
\providecommand{\BIBdecl}{\relax}
\BIBdecl

\bibitem{liu2020deep}
L.~Liu, W.~Ouyang, X.~Wang, P.~Fieguth, J.~Chen, X.~Liu, and
  M.~Pietik{\"a}inen, ``{Deep learning for generic object detection: A
  survey},'' \emph{IJCV}, vol. 128, no.~2, pp. 261--318, 2020.

\bibitem{masi2018deep}
I.~Masi, Y.~Wu, T.~Hassner, and P.~Natarajan, ``Deep face recognition: A
  survey,'' in \emph{SIBGRAPI}.\hskip 1em plus 0.5em minus 0.4em\relax IEEE,
  2018, pp. 471--478.

\bibitem{zheng2016person}
L.~Zheng, Y.~Yang, and A.~G. Hauptmann, ``{Person re-identification: Past,
  present and future},'' \emph{arXiv preprint arXiv:1610.02984}, 2016.

\bibitem{chen2017multi}
X.~Chen, H.~Ma, J.~Wan, B.~Li, and T.~Xia, ``Multi-view 3d object detection
  network for autonomous driving,'' in \emph{CVPR}, 2017, pp. 1907--1915.

\bibitem{mckinney2020international}
S.~M. McKinney, M.~Sieniek, V.~Godbole, J.~Godwin, N.~Antropova, H.~Ashrafian,
  T.~Back, M.~Chesus, G.~C. Corrado, A.~Darzi \emph{et~al.}, ``{International
  evaluation of an AI system for breast cancer screening},'' \emph{Nature},
  vol. 577, no. 7788, pp. 89--94, 2020.

\bibitem{ren2015faster}
S.~Ren, K.~He, R.~Girshick, and J.~Sun, ``{Faster R-CNN: Towards real-time
  object detection with region proposal networks},'' in \emph{NeurIPS}, 2015,
  pp. 91--99.

\bibitem{lin2017focal}
T.-Y. Lin, P.~Goyal, R.~Girshick, K.~He, and P.~Doll{\'a}r, ``Focal loss for
  dense object detection,'' in \emph{ICCV}, 2017, pp. 2980--2988.

\bibitem{he2017mask}
K.~He, G.~Gkioxari, P.~Doll{\'a}r, and R.~Girshick, ``{Mask R-CNN},'' in
  \emph{ICCV}, 2017, pp. 2961--2969.

\bibitem{redmon2016you}
J.~Redmon, S.~Divvala, R.~Girshick, and A.~Farhadi, ``{You only look once:
  Unified, real-time object detection},'' in \emph{CVPR}, 2016, pp. 779--788.

\bibitem{liu2016ssd}
W.~Liu, D.~Anguelov, D.~Erhan, C.~Szegedy, S.~Reed, C.-Y. Fu, and A.~C. Berg,
  ``{SSD: Single shot multibox detector},'' in \emph{ECCV}.\hskip 1em plus
  0.5em minus 0.4em\relax Springer, 2016, pp. 21--37.

\bibitem{law2018cornernet}
H.~Law and J.~Deng, ``{Cornernet: Detecting objects as paired keypoints},'' in
  \emph{ECCV}, 2018, pp. 734--750.

\bibitem{cai2018cascade}
Z.~Cai and N.~Vasconcelos, ``{Cascade R-CNN: Delving into high quality object
  detection},'' in \emph{CVPR}, 2018, pp. 6154--6162.

\bibitem{everingham2010pascal}
M.~Everingham, L.~Van~Gool, C.~K. Williams, J.~Winn, and A.~Zisserman, ``The
  pascal visual object classes (voc) challenge,'' \emph{IJCV}, vol.~88, no.~2,
  pp. 303--338, 2010.

\bibitem{lin2014microsoft}
T.-Y. Lin, M.~Maire, S.~Belongie, J.~Hays, P.~Perona, D.~Ramanan,
  P.~Doll{\'a}r, and C.~L. Zitnick, ``{Microsoft COCO: Common objects in
  context},'' in \emph{ECCV}.\hskip 1em plus 0.5em minus 0.4em\relax Springer,
  2014, pp. 740--755.

\bibitem{ILSVRC15}
O.~Russakovsky, J.~Deng, H.~Su, J.~Krause, S.~Satheesh, S.~Ma, Z.~Huang,
  A.~Karpathy, A.~Khosla, M.~Bernstein, A.~C. Berg, and L.~Fei-Fei, ``{ImageNet
  Large Scale Visual Recognition Challenge},'' \emph{IJCV}, vol. 115, no.~3,
  pp. 211--252, 2015.

\bibitem{DBLP:journals/corr/abs-1811-00982}
A.~Kuznetsova, H.~Rom, N.~Alldrin, J.~Uijlings, I.~Krasin, J.~Pont-Tuset,
  S.~Kamali, S.~Popov, M.~Malloci, T.~Duerig \emph{et~al.}, ``{The Open Images
  Dataset V4: Unified image classification, object detection, and visual
  relationship detection at scale},'' \emph{IJCV}, vol. 128, pp. 1956--1981,
  2020.

\bibitem{tarvainen2017mean}
A.~Tarvainen and H.~Valpola, ``{Mean teachers are better role models:
  Weight-averaged consistency targets improve semi-supervised deep learning
  results},'' in \emph{NeurIPS}, 2017, pp. 1195--1204.

\bibitem{french2017self}
G.~French, M.~Mackiewicz, and M.~Fisher, ``Self-ensembling for visual domain
  adaptation,'' no.~6, 2018.

\bibitem{jeong2019consistency}
J.~Jeong, S.~Lee, J.~Kim, and N.~Kwak, ``Consistency-based semi-supervised
  learning for object detection,'' in \emph{NeurIPS}, 2019, pp.
  10\,758--10\,767.

\bibitem{radosavovic2018data}
I.~Radosavovic, P.~Doll{\'a}r, R.~Girshick, G.~Gkioxari, and K.~He, ``{Data
  distillation: Towards omni-supervised learning},'' in \emph{CVPR}, 2018, pp.
  4119--4128.

\bibitem{he2019momentum}
K.~He, H.~Fan, Y.~Wu, S.~Xie, and R.~Girshick, ``Momentum contrast for
  unsupervised visual representation learning,'' pp. 9729--9738, 2020.

\bibitem{kolesnikov2019revisiting}
A.~Kolesnikov, X.~Zhai, and L.~Beyer, ``Revisiting self-supervised visual
  representation learning,'' in \emph{CVPR}, 2019, pp. 1920--1929.

\bibitem{goyal2019scaling}
P.~Goyal, D.~Mahajan, A.~Gupta, and I.~Misra, ``Scaling and benchmarking
  self-supervised visual representation learning,'' in \emph{ICCV}, 2019, pp.
  6391--6400.

\bibitem{doersch2017multi}
C.~Doersch and A.~Zisserman, ``Multi-task self-supervised visual learning,'' in
  \emph{ICCV}, 2017, pp. 2051--2060.

\bibitem{oquab2015object}
M.~Oquab, L.~Bottou, I.~Laptev, and J.~Sivic, ``{Is object localization for
  free?-Weakly-supervised learning with convolutional neural networks},'' in
  \emph{CVPR}, 2015, pp. 685--694.

\bibitem{wan2018min}
F.~Wan, P.~Wei, J.~Jiao, Z.~Han, and Q.~Ye, ``Min-entropy latent model for
  weakly supervised object detection,'' in \emph{CVPR}, 2018, pp. 1297--1306.

\bibitem{zhang2019leveraging}
D.~Zhang, J.~Han, L.~Zhao, and D.~Meng, ``Leveraging prior-knowledge for weakly
  supervised object detection under a collaborative self-paced curriculum
  learning framework,'' \emph{IJCV}, vol. 127, no.~4, pp. 363--380, 2019.

\bibitem{zhu2005semi}
X.~J. Zhu, ``Semi-supervised learning literature survey,'' University of
  Wisconsin-Madison Department of Computer Sciences, Tech. Rep., 2005.

\bibitem{zhu2009introduction}
X.~Zhu and A.~B. Goldberg, ``Introduction to semi-supervised learning,''
  \emph{Synthesis lectures on artificial intelligence and machine learning},
  vol.~3, no.~1, pp. 1--130, 2009.

\bibitem{chapelle2009semi}
O.~Chapelle, B.~Scholkopf, and A.~Zien, ``Semi-supervised learning (chapelle,
  o. et al., eds.; 2006)[book reviews],'' \emph{IEEE Trans. Neural Netw.},
  vol.~20, no.~3, pp. 542--542, 2009.

\bibitem{lee2013pseudo}
D.-H. Lee, ``{Pseudo-label: The simple and efficient semi-supervised learning
  method for deep neural networks},'' in \emph{ICML Workshop on challenges in
  representation learning}, vol.~3, 2013, p.~2.

\bibitem{hinton2015distilling}
G.~Hinton, O.~Vinyals, and J.~Dean, ``Distilling the knowledge in a neural
  network,'' \emph{arXiv preprint arXiv:1503.02531}, 2015.

\bibitem{phuong2019towards}
M.~Phuong and C.~Lampert, ``Towards understanding knowledge distillation,'' in
  \emph{ICML}, 2019, pp. 5142--5151.

\bibitem{rasmus2015semi}
A.~Rasmus, M.~Berglund, M.~Honkala, H.~Valpola, and T.~Raiko, ``Semi-supervised
  learning with ladder networks,'' in \emph{NeurIPS}, 2015, pp. 3546--3554.

\bibitem{sajjadi2016regularization}
M.~Sajjadi, M.~Javanmardi, and T.~Tasdizen, ``Regularization with stochastic
  transformations and perturbations for deep semi-supervised learning,'' in
  \emph{NeurIPS}, 2016, pp. 1163--1171.

\bibitem{laine2016temporal}
S.~Laine and T.~Aila, ``Temporal ensembling for semi-supervised learning,''
  \emph{arXiv preprint arXiv:1610.02242}, 2016.

\bibitem{tang2020proposal}
P.~Tang, C.~Ramaiah, R.~Xu, and C.~Xiong, ``Proposal learning for
  semi-supervised object detection,'' \emph{arXiv preprint arXiv:2001.05086},
  2020.

\bibitem{viola2001rapid}
P.~Viola and M.~Jones, ``Rapid object detection using a boosted cascade of
  simple features,'' in \emph{CVPR}, vol.~1, 2001, pp. I--I.

\bibitem{girshick2014rich}
R.~Girshick, J.~Donahue, T.~Darrell, and J.~Malik, ``Rich feature hierarchies
  for accurate object detection and semantic segmentation,'' in \emph{CVPR},
  2014, pp. 580--587.

\bibitem{girshick2015fast}
R.~Girshick, ``{Fast R-CNN},'' in \emph{ICCV}, 2015, pp. 1440--1448.

\bibitem{lin2017feature}
T.-Y. Lin, P.~Doll{\'a}r, R.~Girshick, K.~He, B.~Hariharan, and S.~Belongie,
  ``Feature pyramid networks for object detection,'' in \emph{CVPR}, 2017, pp.
  2117--2125.

\bibitem{huang2015densebox}
L.~Huang, Y.~Yang, Y.~Deng, and Y.~Yu, ``{Densebox: Unifying landmark
  localization with end to end object detection},'' \emph{arXiv preprint
  arXiv:1509.04874}, 2015.

\bibitem{redmon2017yolo9000}
J.~Redmon and A.~Farhadi, ``{YOLO9000: better, faster, stronger},'' in
  \emph{CVPR}, 2017, pp. 7263--7271.

\bibitem{redmon2018yolov3}
J.~Redmon and A.~Farhadi, ``{YOLOv3: An incremental improvement},'' \emph{arXiv
  preprint arXiv:1804.02767}, 2018.

\bibitem{zhou2019bottom}
X.~Zhou, J.~Zhuo, and P.~Krahenbuhl, ``Bottom-up object detection by grouping
  extreme and center points,'' in \emph{CVPR}, 2019, pp. 850--859.

\bibitem{zhu2019feature}
C.~Zhu, Y.~He, and M.~Savvides, ``Feature selective anchor-free module for
  single-shot object detection,'' in \emph{CVPR}, 2019, pp. 840--849.

\bibitem{duan2019centernet}
K.~Duan, S.~Bai, L.~Xie, H.~Qi, Q.~Huang, and Q.~Tian, ``{Centernet: Keypoint
  triplets for object detection},'' in \emph{ICCV}, 2019, pp. 6569--6578.

\bibitem{russakovsky2015imagenet}
O.~Russakovsky, J.~Deng, H.~Su, J.~Krause, S.~Satheesh, S.~Ma, Z.~Huang,
  A.~Karpathy, A.~Khosla, M.~Bernstein \emph{et~al.}, ``Imagenet large scale
  visual recognition challenge,'' \emph{IJCV}, vol. 115, no.~3, pp. 211--252,
  2015.

\bibitem{he2016deep}
K.~He, X.~Zhang, S.~Ren, and J.~Sun, ``Deep residual learning for image
  recognition,'' in \emph{CVPR}, 2016, pp. 770--778.

\bibitem{hu2018squeeze}
J.~Hu, L.~Shen, and G.~Sun, ``Squeeze-and-excitation networks,'' in
  \emph{CVPR}, 2018, pp. 7132--7141.

\bibitem{beluch2018power}
W.~H. Beluch, T.~Genewein, A.~N{\"u}rnberger, and J.~M. K{\"o}hler, ``The power
  of ensembles for active learning in image classification,'' in \emph{CVPR},
  2018, pp. 9368--9377.

\bibitem{reed2014training}
S.~Reed, H.~Lee, D.~Anguelov, C.~Szegedy, D.~Erhan, and A.~Rabinovich,
  ``Training deep neural networks on noisy labels with bootstrapping,''
  \emph{arXiv preprint arXiv:1412.6596}, 2014.

\bibitem{dai2016r}
J.~Dai, Y.~Li, K.~He, and J.~Sun, ``{R-FCN: Object detection via region-based
  fully convolutional networks},'' in \emph{NeurIPS}, 2016, pp. 379--387.

\bibitem{pytorch_faster}
\BIBentryALTinterwordspacing
FAIR. (2018) {Faster R-CNN and Retina network in PyTorch 1.0: Model zoo and
  baselines}. [Online]. Available:
  \url{https://github.com/facebookresearch/maskrcnn-benchmark}
\BIBentrySTDinterwordspacing

\bibitem{bell2016inside}
S.~Bell, C.~Lawrence~Zitnick, K.~Bala, and R.~Girshick, ``{Inside-outside net:
  Detecting objects in context with skip pooling and recurrent neural
  networks},'' in \emph{CVPR}, 2016, pp. 2874--2883.

\end{thebibliography}

\end{document}